\documentclass[11pt,a4paper]{article}
\usepackage[hyperref]{emnlp2020}
\usepackage[ruled,vlined]{algorithm2e}
\usepackage{amsthm}
\usepackage{times}
\usepackage{latexsym}
\usepackage{multicol}
\usepackage{amssymb}
\usepackage{amsmath}
\usepackage{graphicx}
\usepackage[frozencache, cachedir=cache]{minted}

\definecolor{darkgreen}{rgb}{0.0, 0.2, 0.13}

\usepackage{microtype}
\usepackage{graphicx}
\aclfinalcopy 
\def\aclpaperid{3495} 

\title{Neural Conversational QA: Learning to Reason vs Exploiting Patterns}

\author{
    Nikhil Verma$^1$ \quad
    Abhishek Sharma$^{1,2}$\thanks{\,  Work done during internship at IBM Research AI} \quad
    Dhiraj Madan$^1$ \\
    {\bf Danish Contractor}$^1$ \quad 
    {\bf Harshit Kumar}$^1$ \quad 
    {\bf Sachindra Joshi}$^1$ \\
    $^1$IBM Research AI, New Delhi \qquad  
    $^2$Indian Institute of Technology (BHU), Varanasi \\
    \texttt{nikhilweee@gmail.com,abhishek.sharma.mat16@iitbhu.ac.in} \\
    \texttt{\{dmadan07,dcontrac,harshitk,jsachind\}@in.ibm.com}
}

\date{}

\begin{document}
\maketitle

\begin{abstract}
Neural Conversational QA tasks like ShARC require systems to answer questions based on the contents of a given passage. On studying recent state-of-the-art models on the ShARC QA task, we found indications that the models learn spurious clues/patterns in the dataset. Furthermore, we show that a heuristic-based program designed to exploit these patterns can have performance comparable to that of the neural models. In this paper we share our findings about four types of patterns found in the ShARC corpus and describe how neural models exploit them. Motivated by the aforementioned findings, we create and share a modified dataset that has fewer spurious patterns, consequently allowing models to learn better.
\end{abstract}

\section{Introduction} 
\label{sec:intro}

\begin{figure}[ht]
    \small
    \centering
    \includegraphics[scale=0.9]{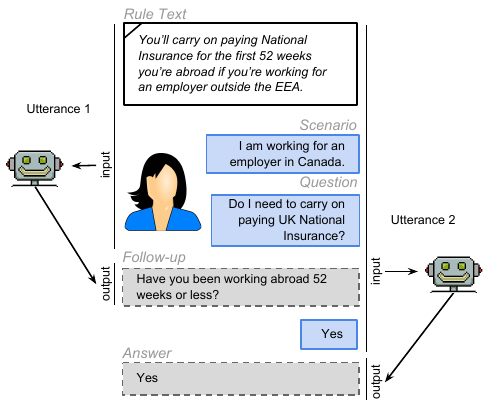}
    \caption{\small{Dialog Flow in ShARC (\cite{ShARC})}}
    \label{fig:cqa}
\end{figure}

ShARC, a conversational QA task \cite{ShARC}, requires a system to answer user questions based on \emph{rules} expressed in natural language text. An example in Figure~\ref{fig:cqa} shows a user sharing some background information (referred to as \emph{scenario}) and asking a question about continuing to pay for `UK National Insurance'. The \emph{rule text} associated with this dialog exchange defines the policy that guides the conversation flow. At any turn in the conversation, a system may choose to respond with a final Yes/No answer; ask a follow-up question to obtain more information from the user; or reply that the question is irrelevant to the context.

Several deep learning models such as BERT-QA \cite{BERT}, E3 \cite{E3}, and BiSon \cite{BiSon} perform reasonably well on this task. However, our exploration of the ShARC dataset indicates that there are multiple spurious patterns that could be exploited by neural models. We observe that the performance of the models mentioned above drops when they are tested on a perturbed dataset, suggesting that the underlying neural models do not generalize and are rather over-sensitive \cite{welbl2020undersensitivity} to minor textual perturbations. By sensitivity we refer to a model's ability to generalize itself but not over-fit, while still being invariant to perturbations or text transformations \cite{teney2020value, szegedy2013intriguing}. Our observations about conversational QA models designed for ShARC learning spurious statistical clues are in line with those reported by \citet{ProbingNN}. To the best of our knowledge, we are the first to demonstrate this problem in conversational QA.

\noindent{\bf Patterns in the ShARC dataset:}
We discover four types of patterns in the ShARC dataset: (1) correlation between the last answer to a follow-up question and the predicted answer to a user question; (2) a high correlation between asking a new follow-up question and the number of turns in the dialog history; (3) correlation between the sequence of follow-up questions in the dialog history and the sequence of rule clauses in the rule-text; and (4) correlation between an empty history/scenario and the answer being irrelevant. 

\noindent{\bf Contributions:} 
The main contributions of this work are as follows: (1) We present a simple heuristics driven program designed to exploit the aforementioned patterns which has performance comparable to the state-of-the-art models. (2) The performance of the state-of-the-art models drops when they are tested on a {\em perturbed} test set that has these patterns diluted or removed. (3) We also identify a weakness in the current evaluation procedure, and propose an improved evaluation criteria which penalizes BLEU scores if a follow-up question is not generated when it should be. We refer to this criteria as BLEU-P (BLEU penalized) in the rest of the paper. (4) We generate a new dataset which reduces the patterns identified in the original dataset, and re-benchmark existing state-of-the-art models published on the leaderboard. We find that the models learn better on this dataset and their performance is consistent across the original and the perturbed dev sets. Our dataset and all accompanying scripts are available at \url{https://github.com/nikhilweee/neural-conv-qa}.

\section{Patterns in the ShARC Dataset}
\label{sec:patterns}

This section describes the spurious patterns in the ShARC dataset and presents a simple heuristic based program designed to exploit these patterns.  

\par \textbf{Pattern 1: Last follow-up answer is the predicted answer:}
Based on the asterisk (`*') as a separator between rule clauses, we found that $54.52\%$ of the instances consist of a list of only-{\em conjunctive} or only-{\em disjunctive} clause conditions. Consider a case where a {\em rule} consists of only conjunctive clause statements. If any single follow-up question, generated based on one of these clauses, is answered with a `No' by a user, the answer to the user's question shall be `No'. In this case, no follow-up questions need to be asked. Thus, one often finds the sequence of follow-up answers in the dialog history as (`Yes', \ldots, `Yes', `No'), for which the answer to the user question is a `No'. Similarly, corresponding to a case where the {\em rule} consists of only disjunctive clauses, the follow-up answer sequence is (`No', \ldots, `No', `Yes'), the answer to which is a `Yes'. This indicates a high correlation between the final answer and the last answer of the history. We found that $74.6\%$ of the instances in the train set with a `Yes'/`No' answer have the same answer as that of the last follow-up question. Although this is reflective of real-world conversations, a model can do a good job on this task by exploiting just this pattern. 

\par \textbf{Pattern 2: Likelihood of asking a follow-up question decreases with number of turns}:
It is intuitive to expect that as the number of follow-up questions that have been asked increases, the likelihood of asking another follow-up question decreases (clauses are finite). 
Appendix \ref{app:pattern-2}, contains an empirical study on the training data, and demonstrates the decrease in the probability of asking a follow-up question with the increase in the number of turns in the dialog history.

\par \textbf{Pattern 3: Follow-up questions occur in the same sequence as the rule clauses in the passage:}
As discussed earlier, many of the rule clauses tend to be conjunctive/disjunctive. Thus, the next follow-up question that one needs to ask is not unique, since one can always consider any of the statements in the clause that has not been considered so far. However, the ground truth data considers these clauses in sequential order to generate the follow-up questions. Among all instances where a conjunctive/disjunctive clause can be discerned and have a follow-up question generated as a part of the ground truth, $62.8\%$ satisfy the condition that the first clause that has not yet been asked is indeed the next follow-up question. We explain this pattern in detail and discuss how it affects computation of the BLEU metric in Appendix \ref{app:pattern-3}. 

\par \textbf{Pattern 4: Answer as `Irrelevant':} Amongst the train instances where user background information and dialog history is empty, $66.67\%$  have the final answer as \textit{Irrelevant}.

\subsection{A Simple Heuristics-based Program}
To demonstrate the ease with which these patterns can be exploited by a model, we create a simple program that follows a set of hand-crafted rules. The program takes the following actions:

\noindent \textbf{1. Answer `Irrelevant':} If the following conditions are jointly satisfied: a) no follow-up questions have been asked so far; b) the background information (scenario) is empty; c) there is low word overlap between the rule and the question; then the program answers \textit{Irrelevant}.

\noindent \textbf{2. Generate `Follow-up Question':} If the previous condition fails and the number of clauses in the rule are more than the number of follow-up questions asked, then a follow-up question is predicted, and the model asks the next clause in the rule text as a question by appending the words ``\emph{Are you}" in the beginning and a question mark at the end.
  
\noindent \textbf{3. Answer with a `Yes' or a `No':} If both the above scenarios are false, then the model response to the user question is the user's response to the last follow-up question.

\section{Evaluation Metrics in ShARC}
The following metrics are used in the evaluation of the ShARC task:

\begin{enumerate}
    \item {\bf Micro and Macro Accuracy:} 
At each turn, the model response is either a \emph{Yes/No/Irrelevant} or a \emph{follow-up} question. The micro and macro accuracy measures the ability of a model to correctly predict these four classes.
    \item {\bf BLEU:} This is used to assess the correctness of the follow-up question generated in case the model chooses to generate one.  
\end{enumerate}

Our experiments with the dataset suggest two weaknesses in the evaluation of follow-up questions which we discuss below.

\noindent \textbf{1. Incomplete reference set:}
Recall that Pattern 3 suggests the existence of a sequential correlation between the rule clauses\footnote{conjunctive-only / disjunctive-only clauses} in the rule text and the follow-up questions. This means that if an out of sequence follow-up question is generated by a model, then it is incorrectly penalized because the evaluation script expects the next follow-up question that would have occured in the original rule sequence. To mitigate this penalization, we create a list of alternative candidate references using the clauses in {\em conjunctive}-only or {\em disjunctive}-only instances in dataset. We make use of the standard implementation of BLEU which supports multiple references \cite{BLEU}. To generate multiple candidate references for the ShARC dataset, we identify instances which have a follow-up question as the gold answer and the follow-up question seems to be based on one of the clause statements in the rule text. We then create alternative follow-up questions from each of the clause statements which have not been a part of any follow-up question in the history. Please see Appendix \ref{app:multi-ref} for details about the algorithm to generate these alternative follow-ups. These alternative follow-up questions constitute a set of candidate questions. In the rest of the paper, we refer to this BLEU score computed using multiple references as \emph{Multi-BLEU}, to distinguish it from the officially reported BLEU metric.

\noindent \textbf{2. Improper Penalization for BLEU:}
As mentioned in the section \ref{sec:intro}, the official evaluation scripts do not penalize BLEU scores of a model if it does not predict a follow-up question, and rather predicts a final answer (\emph{Yes/No} or \emph{Irrelevant}). This is because it considers only those cases where both the ground-truth and the model predictions are \emph{follow-up}s. A hypothetical model which classifies only 1 test instance as follow-up and produces it perfectly can get a BLEU score of 100 in this metric. We therefore update the evaluation script to penalize BLEU score in such cases, and refer to this as BLEU-P (BLEU-Penalized). This considers all instances where the ground truth is a follow-up question. We use the predicted response from the model as the answer to evaluate. This effectively counts a BLEU of 0 in almost all cases for these instances. When we use multiple references for computing BLEU, we refer to this as Multi-BLEU-P.

\section{Evidence of Patterns}
\label{sec:experiments}

Table~\ref{tab:original} and \ref{tab:perturbed-history} report results of our experimental study, providing evidence to support the following: (a) The heuristics-based program has performance comparable to the state-of-the-art models. (b) The performance of models drops when they are trained on the original dev set and tested on the perturbed dev set that has diluted or reduced patterns, indicating a reliance on patterns (c) The official evaluation scripts do not penalize BLEU scores and do not consider other candidate answers in their calculations.

To prepare a perturbed dev set, we modify the official dev set by shuffling the dialog turns of the history of a conversation (We modify approximately $20\%$ of the dev set. For more details, please refer to Appendix \ref{app:statistics}). We refer to this set as the ``History-Shuffled'' dev set, a perturbation introduced by shuffling dialog history to dilute Patterns $1$ \& $3$. Note that shuffling the dialog history introduces examples which are unlikely to occur in real conversations (Eg. Asking a follow-up question based on a set of conjunctive rule-clauses even though a user has already responded with a ``No''). For model details, please refer to Section \ref{sec:recommendations}.

Using the evaluation metrics (micro and macro accuracy for turn level classification; and Multi-BLEU and Multi-BLEU-P for answer generation accuracy), we report the performance of the top two ranked models \footnote{\url{https://sharc-data.github.io/leaderboard.html}} (E3 \cite{E3} and BiSon \cite{BiSon}), the BERT based model (BERT-QA ~\cite{BERT}) and our heuristic based model by training them on the official ShARC training set and evaluating them on the original and History-Shuffled dev sets. We use the code released by respective authors for E3 and BiSon. We also use the same hyperparameters as mentioned in the respective papers.
 
\noindent{\bf Weakness in Offical BLEU scores:} In Table~\ref{tab:original}, the BLEU scores are lower than the Multi-BLEU scores, by an average of $16.03\%$. This is as expected since the official scripts do not account for valid alternatives and the gold answers have been generated in accordance with Pattern 3. Furthermore the models result in significantly lower BLEU scores on the Multi-BLEU-P metric as it penalizes models if they dont generate a follow-up question when they were supposed to. This suggests that the official scripts grossly over-estimate model performance (BiSon's actual Multi-BLEU-P is score is only 14.25). In the rest of the experiments we only report Multi-BLEU and Multi-BLEU-P scores.

\noindent{\bf Heuristic model and effect of Patterns:} Tables \ref{tab:original} and \ref{tab:perturbed-history} show that the heuristic-based model not only has comparable performance across all other models but also across both dev sets (original and History-Shuffled). If we look at the first two columns (micro and macro accuracy), all models tested on the History-Shuffled dev set report a drop in performance. The average drops in micro \& macro accuracies are $8.16\%$ and $5.3\%$ respectively. While changes in performance are can be attributed to change in train and test distributions, the goal of this experiment is to demonstrate that all models are relying on a spurious pattern induced by the sequence of follow-up answers in dialog history. Thus, it is interesting to note that a dilution of just $20\%$ of the patterns leads to a sizeable drop in performance.

\begin{table}
    \small
    \center
    \begin{tabular}{|p{1.2cm}|p{0.7cm}|p{0.7cm}|p{0.7cm}|p{0.7cm}|p{1.1cm}|}
    \hline
    \multicolumn{6}{|c|}{Train and Eval on Original dataset}                           \\ \hline
    Model      & Micro Acc   & Macro Acc   & BLEU        & Multi BLEU  & Multi BLEU P  \\ \hline
    Heuristics & 63.74       & 71.25       & {\bf 47.57} & 52.81       & {\bf 36.90}   \\ \hline
    BERTQA     & {\bf 68.63} & 73.67       & 47.36       & 54.04       & 35.94         \\ \hline
    E3         & 67.63	     & {\bf 73.79} & 46.29	     & {\bf 54.64} & 39.36         \\ \hline
    BiSon      & 65.95       & 70.79       & 46.62       & 54.06       & 14.25         \\ \hline
    \end{tabular}
    \caption{Models trained on the original ShARC training set and tested on the original dev set. The maximum score for every metric is highlighted in bold.}
    \label{tab:original}
\end{table}

\begin{table}
    \small
    \center
    \begin{tabular}{|p{1.2cm}|p{0.7cm}|p{0.7cm}|p{0.7cm}|p{1.1cm}|}
    \hline
    \multicolumn{5}{|c|}{Train on Original, Eval on History-Shuffled}    \\ \hline
    Model      & Micro Acc   & Macro Acc   & Multi BLEU  & Multi BLEU P  \\ \hline
    Heuristics & 58.46       & 67.42       & {\bf 52.81} & {\bf 36.90}   \\ \hline
    BERTQA     & 63.39       & 69.86       & 53.99       & 35.70         \\ \hline
    E3         & 63.52       & 70.07       & 42.63       & 38.70         \\ \hline
    BiSon      & 61.45       & 67.55       & 53.56       & {\bf 14.27}   \\ \hline
    \end{tabular}
    \caption{Models trained on the original ShARC training set and tested on the History-Shuffled dev set. All scores except the ones highlighted in bold suffer a drop when compared to Table \ref{tab:original}.}
    \label{tab:perturbed-history}
\end{table}

\section{Modified ShARC dataset}
\label{sec:sharc-mod}
In an attempt to mitigate the effects of the patterns listed in section \ref{sec:patterns} and to reduce the sensitivity of neural models, we create a modified version of the ShARC dataset. For each occurence of an instance conforming to any of the patterns, we automatically construct alternatives where we choose to either \emph{replace} the current instance with an alternative instance which does not exhibit the pattern; or \emph{retain} the original instance. The alternative instances are generated using pattern-specific modifications. For example, we shuffle dialog history to reduce the effect of Patterns 1 \& 3 (For more details, please see appendices \ref{app:sharc-mod} and \ref{app:statistics}). We individually modify both the official train and dev datasets and refer to them as \texttt{ShARC-Mod}.

\begin{table}
    \small
    \center
    \begin{tabular}{|p{1.2cm}|p{0.7cm}|p{0.7cm}|p{0.7cm}|p{1.1cm}|}
    \hline
    \multicolumn{5}{|c|}{Train and Eval on ShARC-Mod}                    \\ \hline
    Model      & Micro Acc   & Macro Acc   & Multi BLEU  & Multi BLEU P  \\ \hline
    Heuristics & 56.52       & 56.18       & 52.81       & 36.90         \\ \hline
    BERTQA     & 66.04       & 70.88       & 44.32       & 27.14         \\ \hline
    E3         & 62.56       & 69.82       & 49.82       & 44.56         \\ \hline
    BiSon      & 56.61       & 60.96       & 73.54       & 01.28         \\ \hline
    \end{tabular}
    \caption{Models trained on the ShARC-Mod training set and evaluated on the ShARC-Mod dev set.}
    \label{tab:eval-mod}
\end{table}

\begin{table}
    \small
    \center
    \begin{tabular}{|p{1.4cm}|p{0.7cm}|p{0.7cm}|p{0.7cm}|p{1.1cm}|}
    \hline
    \multicolumn{5}{|c|}{Train on ShARC-Mod, Eval on Original}           \\ \hline
    Model      & Micro Acc   & Macro Acc   & Multi BLEU  & Multi BLEU P  \\ \hline
    Heuristic  & 63.74       & 71.25       & 52.81       & 36.90         \\ \hline
    BERT-QA    & 66.52       & 71.66       & 44.32       & 27.14         \\ \hline
    E3         & 62.86       & 70.34       & 49.65       & 35.62         \\ \hline
    BiSon      & 57.31       & 61.93       & 73.54       & 01.28         \\ \hline
    \end{tabular}
    \caption{Models trained on the ShARC-Mod training set and evaluated on the original dev set.}
    \label{tab:eval-orig}
\end{table}

\begin{table}
    \small
    \center
    \begin{tabular}{|p{1.4cm}|p{0.7cm}|p{0.7cm}|p{0.7cm}|p{1.1cm}|}
    \hline
    \multicolumn{5}{|c|}{Train on ShARC-Mod, Eval on History-Shuffled}   \\ \hline
    Model      & Micro Acc   & Macro Acc   & Multi BLEU  & Multi BLEU P  \\ \hline
    Heuristic  & 58.47       & 67.42       & 52.81       & 36.90         \\ \hline
    BERT-QA    & 66.26       & 71.47       & 44.28       & 27.10         \\ \hline
    E3         & 63.22       & 70.58       & 49.49       & 43.97         \\ \hline
    BiSon      & 57.00       & 61.70       & 72.57       & 01.20         \\ \hline
    \end{tabular}
    \caption{Models trained using ShARC-Mod and evaluated on the History-Shuffled dev set}
    \label{tab:eval-history}
\end{table}

\subsection{Benchmarking Experiments}
We train and evaluate all models using the ShARC-Mod train dataset and then test on ShARC-Mod dev set as well as the original dev set (containing all patterns) and the History-Shuffled dev set (containing diluted patterns). Studying tables \ref{tab:eval-mod}, \ref{tab:eval-orig} and \ref{tab:eval-history} shows that all models perform consistently across all dev sets. This suggests that models that were earlier sensitive to perturbations now show consistent performance after being trained on a more robust data set. Note that, except for the heuristic model, this performance is indeed lower than what we had when trained on original data sets. This suggests that the neural models have been merely exploiting patterns in the training data and the performance sharply drops when these cues are absent from training data. The heuristic model has the same numbers as before. This is because the heuristic model is based on rules and is never actually trained. Its performance is indeed invariant to the training data.

\section{Recommendations \& Conclusion}
\label{sec:recommendations}
In this paper we demonstrate how a popular Neural Conversational QA dataset inadvertently encodes patterns. We would like to emphasize that the patterns found, by their very nature, are likely to occur in real world tasks but the same patterns can also cause neural models to learn poorly. We release a modified version of this dataset and also improve evaluation criteria that better reflects model performance. We conclude the paper with a few recommendations for the community.

\noindent{\bf For Dataset creators:}  Patterns may exist in a real-world task and artificially introducing perturbations may be an easy way to help reduce their effects. This may result in `unnatural' instances in the dataset but could help train better models.

\noindent{\bf For Model creators:} (1) Model probing and experimenting with perturbed inputs can give deep insights about how a model is reasoning (2) Experimenting with adversarial inputs early on in the design process can help build better models.

\bibliographystyle{acl_natbib}
\bibliography{references}
\appendix
\section{Appendix}

\subsection{Empirical Study for Pattern 2}
\label{app:pattern-2}

\begin{figure}[ht]
    \center{\includegraphics[width=\columnwidth]{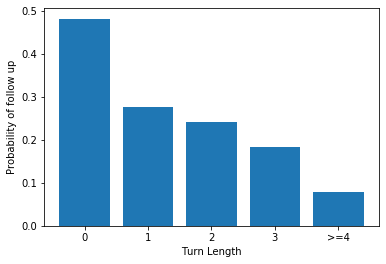}}
    \caption{\label{fig:turns} Probability of asking a follow-up question plotted against the number of turns in the instance.}
    \vspace{-1em}
\end{figure}

Figure \ref{fig:turns} plots the number of turns in an instance vs the probability of asking a follow-up question. It can be seen that with each response, the probability of asking another follow-up question decreases.

\subsection{Detailed explanation for Pattern 3}
\label{app:pattern-3}

Due to the presence of conjunctive/disjunctive clause statements, it becomes ambiguous as to which one to consider for framing the next possible question. As an example, consider the instance described in Table \ref{tab:sample}. The gold answer is a followup question which was framed using the first clause statement. However, follow-up questions framed using any of the other clause statements in the rule such as ``Is the item a lifeboat or an associated equipment, including fuel?", ``Is it medicine or an ingredient for a medicine?", or ``Is it a resuscitation training model?" would have been equally valid. We found that this is a common pattern found in the dataset. More generally, the gold follow-up answer tends to be framed using the first clause which has not been asked so far. To quantify this hypothesis, we followed these steps.

\begin{enumerate}
    \item \label{item:step-1} We filter the instances with a follow-up question as the gold answer and identify clause statements which start with an asterisk (*).
    \item \label{item:step-2} For each follow-up question in the history, we use the longest common substring (LCS) algorithm to compute an intersecting span with the rule text. We then identify the clause statements which intersect with this span. 
    \item \label{item:step-3} We use the same process as above to find a matching clause statement for the follow-up question listed as the answer of the instance.
\end{enumerate}

Amongst the instances identified, we found that $62.8\%$ of them were such that the follow-up question in the answer (step \ref{item:step-3}) intersects with the first clause statement identified in step \ref{item:step-1} that does not appear in the history of the instance.

In an attempt to break this pattern, we identify clause statements in the rule text in the same way as step \ref{item:step-1} above and then shuffle them to create a new instance in \texttt{ShARC-Mod}.

\begin{table}[ht]
\centering
\scriptsize
\begin{tabular}{p{9mm}p{60mm}}
\hline \hline
Rule      & \#\# Items that qualify for the zero rate \\
          & You may be able to apply zero VAT when you sell the \\
          & following to an eligible charity: \\
          &* equipment for making `talking' books and newspapers \\
          &* lifeboats and associated equipment, including fuel \\
          &* medicine or ingredients for medicine \\
          &* resuscitation training models \\
\hline
Scenario  & I used to work for the company, but I quit last month. \\
\hline
Question  & Can I apply zero VAT to this item? \\
\hline
History   &  \\
\hline
Answer    & Is it equipment for making `talking' books and newspapers? \\
\hline \hline
\end{tabular}
\caption{A sample instance from the dev set (utterance \\
id: 0cdee38a5a9cbdda40849861c1edffc1432a3004)} 
\label{tab:sample}
\end{table}

\subsection {Details on creating multiple references}
\label{app:multi-ref}

We discuss the details on how we add additional gold references to the dev dataset. This augmentation only affects the instances which have a follow-up question as the gold answer. If the LCS of the gold answer with the rule text intersects with one of the clause statements (identified in step \ref{item:step-1}), we suspect that other clause statements might also have been one of the possible answers. So we first eliminate the clauses that have already been asked in the history, and again use LCS to find the best matching span for each follow-up question that has been asked in the history so far. If the intersection is with one of the clauses, we eliminate the same. The remaining clauses are then considered potential candidates for the next follow-up question to consider. To create a question from the clause statement, we use a simple heuristic as to finding the words preceeding and following the best match span for the gold answer. These words are then prefixed and suffixed with the other candidate clauses to form potential questions. Algorithm \ref{alg:multi-ref} describes this process in a formal manner. For more details, please refer to the accompanying repository.

\begin{algorithm}[ht]
\SetAlgoLined
 \For{$\text{inst} \in \mathcal{D}$}{
    \If{inst.answer \emph{is} follow-up}{
        Let $\mathcal{C}=\{C_1,C_2,...\}$ be the sequence of clause statements detected ($\mathcal{C}=\emptyset$ if no clause)\;
        
        \emph{span} $\leftarrow$ LCS(inst.gold, inst.rule)\;
        
        \If{$\exists \ i $ \emph{such that} $ C_i \cap \text{span} \neq \emptyset$ }{
            $\mathcal{C}_\text{asked} \leftarrow \{C_j : C_j \cap q \neq \emptyset$ \linebreak[1] $\text{ for some \emph{q} in \emph{inst.follow-up}s}\}$\;
            $\mathcal{C}_\text{candidates} \leftarrow \mathcal{C} \setminus \mathcal{C}_\text{asked}$;
            
            \For{$c \in \mathcal{C}_\emph{candidates}$}{
                Generate follow-up question from $c$ and use it as an additional reference\;
            }
        }{}
    }
 }
\caption{Adding additional references}
\label{alg:multi-ref}
\end{algorithm}

On running this algorithm on the original dev dataset, we were able to add additional references in 183 out of the 2270 instances. It is interesting to note that 96 out of the 183 instances had an empty history, and a follow-up question formed using any of the clause statements in the rule text could have been a valid answer. We also manually evaluate $10\%$ of the generated references and find that barring a few that had minor tense related grammatical issues, all of them were semantically correct.

\subsection{Algorithm for creating \texttt{ShARC-Mod}}
\label{app:sharc-mod}

To create our modified dataset, we perform different modifications depending on whether an instance has scenario or history. Listing \ref{lst:sharc-mod} describes the algorithm to perform these modifications on an instance. More details can be found in the repository accompanying this paper.

\begin{listing}[ht]
\begin{minted}{python}
if not history:
    # history is not present
    if not scenario: 
        if answer in ['yes', 'no']:
            # no history, no scenario,
            # answer is yes/no/irrelevant
            random.choice(shuffle_rules, no_change)
            # deterministic shuffling. 
            # sample from n! - 1 permutations. 
        elif answer in ['irrelevant']:
            random.choice(insert_random_scenario, no_change)
        else:
            # no history, no scenario, answer is span
            # makes sense to ask any question. leave as is.
            pass
    else:
        # scenario present
        if answer in ['yes', 'no', 'irrelevant']:
            # no history, scenario present
            # answer is yes/no/irrelevant
            random.choice(shuffle_rules, no_change)
            # deterministic shuffling. 
            # sample from n! - 1 permutations.             
        else:
            # history present, scenario present
            # answer is span. makes sense to
            # ask any question. leave as is.
            # (this case does not exist)
            pass
else:
    # history is present
    if not scenario:
        if answer in ['yes', 'no', 'more']:
            # history present, no scenario
            # answer is yes/no/irrelevant
            random.choice(shuffle_rule, shuffle_history,
                          shuffle_both, no_change)
            # deterministic shuffling. 
            # sample from n! - 1 permutations. 
        else:
            # history present, no scenario
            # answer is irrelevant
            # (this case does not exist)
            pass

    else:
        # scenario present
        if answer in ['yes', 'no', 'more']:
            # history present, scenario present
            # answer is yes/no/span
            random.choice(shuffle_rule, shuffle_history, 
                          shuffle_both, no_change)
            # deterministic shuffling. 
            # sample from n! - 1 permutations. 
        else:
            # history present, scenario present
            # answer is irrelevant
            # Assert that case does not exist.
            pass
\end{minted}
    \caption{Algorithm for creating \texttt{ShARC-Mod}}
    \label{lst:sharc-mod}
\end{listing}

\subsection{Statistics on the modified datasets}
\label{app:statistics}
In this section, we present some statistics on our modified datasets. 

\textbf{History-Shuffled} dataset: For every instance in the original dataset having more than one questions in its \emph{history}, we either retain or shuffle the order of questions, both with equal probability. This leads to a modification of 5512 out of 21890 instances in the training dataset and 468 out of 2270 instances in the dev dataset.

\textbf{ShARC-Mod} dataset: Using the algorithm in \ref{app:sharc-mod}, out of 21890 training instances, 3287 have the order of \textit{history} shuffled, 3202 have the order of \textit{rules} shuffled, and 2903 instances have both, \textit{history} and \textit{rule} shuffled. Moreover, 596 instances have a random \textit{scenario} added to them. For the dev dataset, out of 2270 instances, 340 have the order of \textit{history} shuffled, 316 have the order of \textit{rules} shuffled, and 323 instances have both, \textit{history} and \textit{rule} shuffled. In this case, 66 instances have a random \textit{scenario} added to them. More details are listed in Appendix \ref{app:sharc-mod}.

\end{document}